\documentclass[12pt,english]{article}

\usepackage[utf8]{inputenc}
\usepackage[english]{babel}
\usepackage{lmodern}
\usepackage[margin={1.5cm,2cm}]{geometry}
\usepackage{multicol}

\usepackage{hyperref}
\usepackage{etoolbox}

\usepackage{graphicx}
\usepackage{subcaption}

\usepackage{booktabs,makecell}
    
\usepackage{siunitx}

\usepackage{amsmath}
\DeclareMathOperator*{\argmax}{\arg\!\max}

\begin{document}
\setlength{\columnsep}{0.5cm}

\title{%
  Exploring BERT Parameter Efficiency on the Stanford Question Answering Dataset v2.0}
\author{
  \textbf{Eric Hulburd}\\
  \small{School of Information}\\
  \small{UC Berkeley}\\
  \small{ehulburd@ischool.berkeley.edu}
}
\date{}
\maketitle

\begin{multicols}{2}
\section*{Abstract}
In this paper we explore the parameter efficiency of BERT \cite{devlin2018bert} on version 2.0 of the Stanford Question Answering dataset (SQuAD2.0). We evaluate the parameter efficiency of BERT while freezing a varying number of final transformer layers as well as including the adapter layers proposed in  \cite{houlsby2019parameterefficient}. Additionally, we experiment with the use of context-aware convolutional (CACNN) filters, as described in \cite{shen2017learning}, as a final augmentation layer for the SQuAD2.0 tasks.

This exploration is motivated in part by \cite{schwartz2019green}, which made a compelling case for broadening the evaluation criteria of artificial intelligence models to include various measures of resource efficiency. While we do not evaluate these models based on their floating point operation efficiency as proposed in \cite{schwartz2019green}, we examine efficiency with respect to training time, inference time, and total number of model parameters. Our results largely corroborate those of \cite{houlsby2019parameterefficient} for adapter modules, while also demonstrating that gains in F1 score from adding context-aware convolutional filters are not practical due to the increase in training and inference time.

\section{Introduction}
Our experiments focus on v2.0 of the SQuAD dataset. The objective of the task is, given an input sequence and a query, to decide whether the answer to the query exists in the input sequence and, if so, where the answer resides within the input sequence. For example, given the below paragraph:

\begin{quote}
In July 2002, Beyonce continued her acting career playing Foxxy Cleopatra alongside Mike Myers in the comedy film, Austin Powers in Goldmember... Beyonce released "Work It Out" as the lead single from its soundtrack album which entered the top ten in the UK, Norway, and Belgium. In 2003, Beyonce starred opposite Cuba Gooding, Jr....
\end{quote}

and query \textit{Who did Beyonce star with in the movie, "Austin Powers in Goldmember"?}, an accurate model would output the indices [84, 85] corresponding to the subsequence "Mike Myers". The objective of the model is thus to output two sequences of logits, one for the start index and one for the end index, corresponding to each word in the input sequence. The model prediction for the query answer is thus given by:

\[i = \argmax_{x \in \mathcal{X}} F(\mathcal{X}, q, \theta)\]

where \textit{i} is the answer index, \(F\) represents the model, \(\theta\) represents its parameters, \(\mathcal{X}\) is the input sequence of words and \textit{q} is the query sequence. An answer of indices [0, 0] indicates that the input sequence does not contain an answer to the query.

Models are scored based on the exact match of indices across the test dataset and the F1 score of tokens correctly identified within the range of the answer indices. Progress on the task has been rapid. The \textit{Allen Institute for Artificial Intelligence} reported a top F1 score of 66.251\% on May 30, 2018 using a BiDAF model \cite{DBLP:journals/corr/SeoKFH16} with self-attention on top of ELMo embeddings. As of this writing, the top score F1 score of 92.425\% was reported on November 6, 2019 using an ALBERT base (discussed below) \cite{squadexplorer}.

\subsection{Related Work}

Here we are interested specifically in relative parameter, training, and inference efficiency under fixed hardware constraints. There have been several encouraging signs of interest in examining models through this perspective. \cite{houlsby2019parameterefficient} examined the efficacy of transfer learning with BERT. We discuss model details in the ensuing section, but here we mention they share a base natural language model (ie BERT) among a diverse set of NLP tasks by injecting adapter layers between each transformer layer, rather than fine tuning the underlying transformer layers themselves. Note, however, that the task specific, adapted models still train layer normalization parameters for all transformer layers. The intention here is to yield compact and extensible downstream models. The authors ran their models successfully on the SQuAD1.0 task (similar to SQuAD2.0 except that it does not ask the models to identify unanswerable questions), approaching full fine-tuning performance with only 2\% of the trainable parameters (they achieve an F1 score on SQuAD1.0 of 90.4\%, while full fine tuning achieves a 90.7\% F1 score). Subsequently, \cite{semnani2019berta} applied additional layers and data augmentation to the adapter-BERT model to the SQuAD2.0 task, achieving an F1 score of 74.7\% (their baseline fine tuned model achieved 76.5\%) with 0.57\% of the number trainable parameters.

Most impressively, \cite{lan2019albert} created a \textit{lite} version of BERT (namely, ALBERT). They were able to improve upon a fully fine tuned BERT-large model on the SQuAD2.0 task with a configuration that had 18x fewer parameters. They achieved this feat with three innovations. First, they separated the word embeddings and the hidden context layers, which allowed them to reduce the word embedding dimension, and thereby the total number of model parameters, without losing richness of context. Second, they shared attention and feed-forward weights between transformer layers. Last, they trained the model on a sentence order prediction task rather than next sentence prediction. Because we did not implement these innovations here, we suggest that our results here regarding adapter layers, layer freezing, and the use of context-aware convolutional filters may extend to ALBERT pending experimental verification.

\section{Models}

\subsection{BERT}

The BERT (Bidirectional Encoder Representations from Transformers) model \cite{devlin2018bert} differentiated itself in couple of ways from existing approaches to natural language understanding. \textit{Global Vectors for Word Representation (GloVe)} \cite{pennington-etal-2014-glove} presented a general approach to sharing vectorized representations of words. The shortcoming of this approach is that words do not uniquely map to meaning. For instance, a robust language model would differentiate "bat" in "the \textit{bat} flow out of the cave" and "the shortstop swung the \textit{bat}".  Subsequently, \cite{DBLP:journals/corr/abs-1802-05365} provided a method to represent meaning within specific contexts, which they termed \textit{Embedding from Language Models (ELMo)}, by using sequences of bidirectional \textit{long short-term memory (LSTM)} cells.

BERT innovated upon existing models by training bi-directionally, and simultaneously rather than sequentially, and by focusing on training generalized language models that could be fine tuned for downstream tasks. The complete architecture consists of a configurable number of multi-headed attention layers as described in \cite{DBLP:journals/corr/VaswaniSPUJGKP17}. The bi-directional nature of the attention mechanism  was leveraged by training the model to identify masked words and predict next sentences from their training corpus. The attention mechanism, called a transformer, matches key values to query values for every token in a sequence and outputting a weighted combination of attention values. For a high level summary of tranformer layers, we refer the reader to the visualizations in \cite{googleaiblogtransformer}.

Once the generalized model has been trained, the model can be re-initialized to output a sequence of contextualized vectors for a given input sequence, which in turn can be passed through an arbitrary final layer to produce outputs for a given task. In \cite{devlin2018bert}, researchers added a single affine layer to produce logits for SQuAD2.0 sequences and fine tuned the underlying BERT weights to achieve an F1 score of 83.1\%, which represented an improvement of 5.1\% over the previous best system.

\subsection{Adapter-BERT}

Motivated to create an architecture for compact and extensible models based on BERT, \cite{houlsby2019parameterefficient} propose an adapter method for training BERT for downstream tasks. For every transformer layer within BERT, \cite{houlsby2019parameterefficient} insert adapter modules after each feed-forward layer. These adapter modules are simply feed-forward layers that project the transformer layer down to an adapter size and subsequently back up to the hidden size of the BERT layers. Each of these adapter modules are followed by layer normalization, which acts to stabilize gradients and improve training efficiency as described in \cite{ba2016layer}. During training on downstream tasks, they freeze the original layers of the BERT model, except for the layer normalization layer and the adapter modules. The authors show performance approaching full fine-tuned BERT models with two orders of magnitude fewer trainable parameters on several tasks, including SQuAD1.0. \cite{semnani2019berta} apply this work successfully to the SQuAD2.0 task with additional improvements in the final layer architecture and training data augmentation.

\subsection{Context Aware Convolutional Filters}

With an eye toward parameter efficiency, we explore the potential of context-aware convolutional filters as a final layer to process BERT output and calculate logits for the SQuAD2.0 task. Originally, \cite{shen2017learning} demonstrated state of the art performance among CNN models on the WikiQA and SelQA datasets in 2017 while employing a context aware CNN model. At a high level, the architecture consists of two convolutional modules, one of which generates a separate set of filters for each input example. The second module applies the feature maps resulting from the first module over the original input as a convolution to produce a second and final set of feature maps.

\cite{shen2017learning} train separate modules for question and answer sequences and use a matching module to calculate logits in a model schema they describe as \textit{Adaptive Question Answering {AdaQA}}. We omit this technique from our analysis for two reasons. First, both BERT and each CACNN require fixed length inputs. This in turn requires that we feed question and answer separately through the BERT model or split the BERT model at question-answer a boundary. The former approach would require fine-tuning BERT twice, once for question and once for answers, which we take to be training inefficient. Worse, it would cut off transformer attention between question and answer. The latter approach is less flawed, but it would require imposing additional restrictions on text and query length, which would be another source of training efficiency. That said, we did not test these assumptions and they remain to be fully corroborated.

In \cite{shen2017learning}, before generating the per sample convolutional filters, the results of the first convolutional module are summed across the length of the input. The result is a single context vector for each input example, which captures context across the length of the input sequence. A separate convolution operation, followed by a concatenation or tile operation, project the context vector up to proper dimensions for the contextualized per sample filters. These are then convolved over the input sequence to produce feature maps, which can be concatenated and passed through a final affine layer in order to produce logits. See \textit{Figure 1}.

We additionally explore a simpler variation of the CACNN architecture. We omit the generation of a context vector and directly use feature maps from the first CNN module to concatenate a series of per sample contextualized filters. See \textit{Figure 2}.

\end{multicols}
\begin{figure}[t!]
  \includegraphics[width=\linewidth]{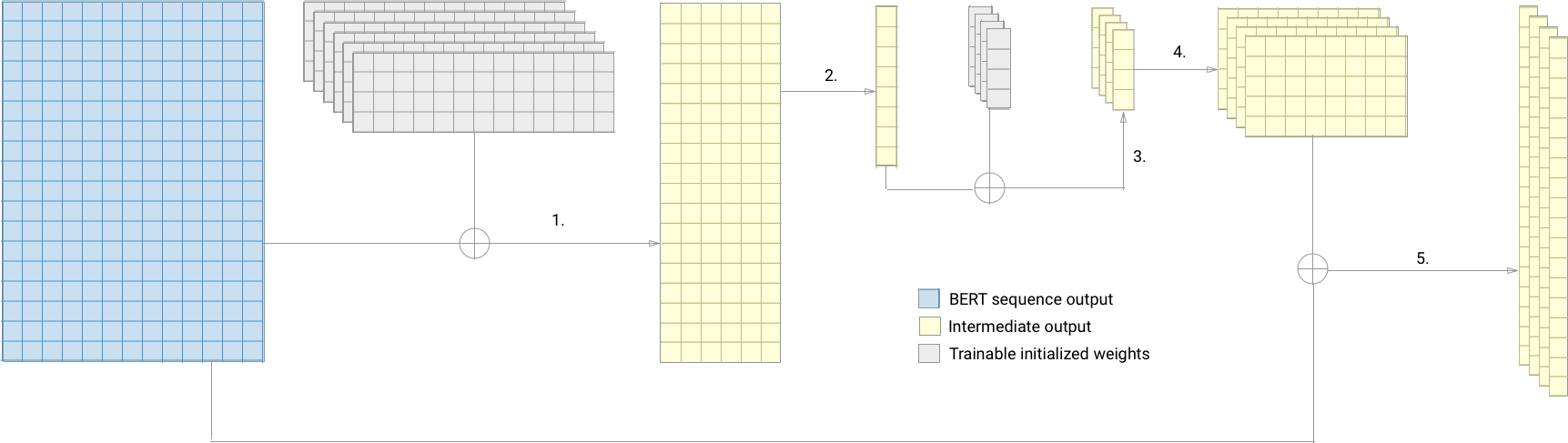}
  \caption{Context-aware CNN with context vectorization module based on \cite{shen2017learning}. 1) Initial convolution on BERT sequence output. Results are concatenated to a two dimensional tensor. 2) Two dimensional tensor reduced along sequence length by taking max, yielding a single context vector. 3) Convolution on context vector, 4) results are concatenated or tiled until appropriate filter size for BERT output sequence. 5) Context aware filters are convolved with BERT sequence output to produce a series of final feature maps.}
\end{figure}
\begin{multicols}{2}

\end{multicols}
\begin{figure}[b!]
  \includegraphics[width=\linewidth]{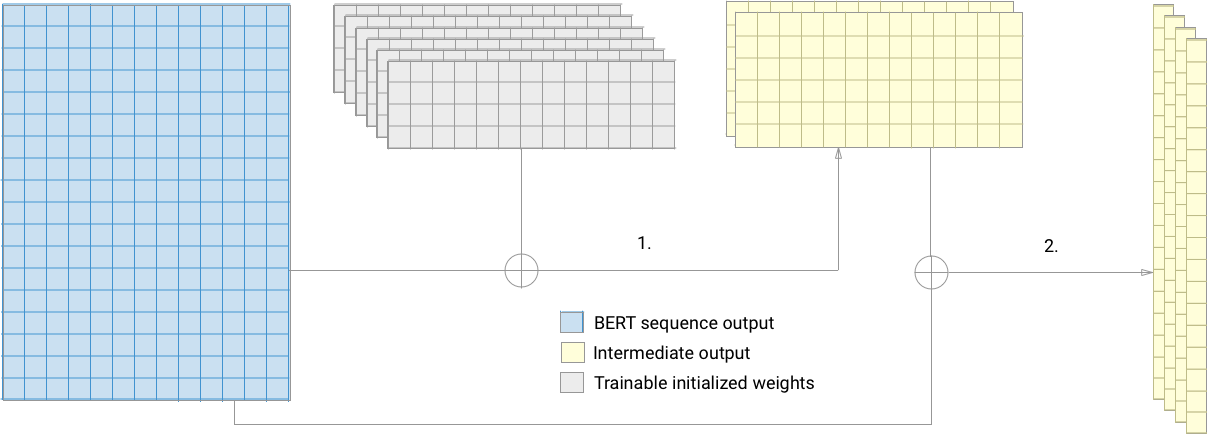}
  \caption{A simplified context-aware CNN module without context vectorization based on \cite{shen2017learning}. 1) Initial convolution on BERT sequence output. Results are concatenated and split into two dimensional tensors appropriate to serve as filters against original BERT sequence output. 2) Context aware filters are convolved with BERT sequence output to produce a series of final feature maps.}
\end{figure}
\begin{multicols}{2}

\section{Experiments and Results}

All of our models were trained on a Lambda Labs Deep Learning Workstation using one of two Titan RTX 24GB GPUs. We use training and prediction batch sizes of eight. Our models are initialized with BERT base checkpoints made available by \cite{devlin2018bert}. All reported training and inference times apply to the complete SQuAD2.0 training and test data using a max sequence length of 384 (the default used by \cite{devlin2018bert}); this yielded 132,300 training examples and 12,320 test examples. Models are trained over three epochs. All of our code modifications were committed to \cite{ehulburdadapterbert}, which is fork of \cite{houlsby2019parameterefficient}.

To establish a baseline, we train the final 0, 1, 3, 6, and 12 transformer layers of the BERT model. Note, even when attention and feed-forward layers are frozen, the layer norms across all transformer layers and final affine layer are trainable. Subsequently, we train BERT with adapter sizes of 64 and 768, as used in \cite{houlsby2019parameterefficient} and \cite{semnani2019berta}. Last, we explore different initial filter configurations and context-vectorized vs. simplified CACNN architectures. We compare results based on F1 score with respect to the total number of trained parameters, training time, and inference time as shown in Appendix I. We compare augmented models with variable final transformer layer freezing to assess whether their gains over a frozen BERT model are indicative of gains when fully fine tuning the BERT base simultaneously.

\subsection{BERT with Layer Freezing}

Our results are summarized in \textit{Table 1}. Training without any layer freezing produces the highest F1 score (76.3), but we note that freezing the bottom six layers (F1 score of 75.2) yields a similar improvement in F1 score with respect to the number of trainable parameters:

$$\dfrac{F1 - 50}{log(N\_trainable\_parameters)} = 3.3$$

This suggests that fine tuning may proceed more easily on some tasks when the bottom layers are not fine tuned (see Appendix 1c). It is similarly optimal when viewed in terms of reduction in training speed (see Appendix 1a). Of course, freezing BERT layers has no impact on either the total number of parameters, nor the inference time. All of the below models completed the inference batch set in 145-150 seconds.

\end{multicols}
\begin{table}[h!]
    \centering
    \label{tab:table-1}
\begin{tabular*}{\textwidth}{c @{\extracolsep{\fill}} @{}*{6}{c}@{}}
    \toprule
      \thead{Model}
        & \thead{No. Layers\\Trained}
        & \thead{No. Trainable\\Params}
        & \thead{EM}
        & \thead{F1}
        & \thead{Train\\Time (min)}\\
    \midrule
      L12 & 12 & 108,311,810 & 73.3 & 76.3 & 255\\
      L6 & 6 & 42,548,738 & 72.3 & 75.2 & 218\\
      L3 & 3 & 21,294,338 & 69.3 & 72.5 & 203\\
      L1  & 1 & 7,124,738 & 62.8 & 65.9 & 193\\
      L0 & 0 & 39,938 & 52.1 & 53.1 & 188\\
    \bottomrule
\end{tabular*}
\caption{SQuAD2.0 results with variable transformer layer freezing. Note, regardless of a layer frozen state, layer normalization and final affine layers are trained in all of the above models.}
\end{table}
\begin{multicols}{2}

\subsection{Adapter Modules}

We run BERT with adapter sizes of 64 as used in \cite{houlsby2019parameterefficient} for SQuAD1.0 task and 768 as used in \cite{semnani2019berta}. In contrast to \cite{semnani2019berta}, we do not make any SQuAD2.0 specific changes to the model for evaluating the adapter approach. Our results confirm the efficacy of the adapter approach for fine tuning. The 768 sized adapter model with all BERT layers frozen comes very close to the full fine-tuned BERT model - 75.0\% F1 compared to 75.8\ - while using about 26\% of the number of parameters (see Appendix 1c). Note, however, that the 768 adapter size model requires no less training time on a single GPU as the fine tuned BERT models (see Appendix 1a) and adds over thirty seconds, or about 20\%, to the inference time (see Appendix 1b). Given the inference dataset of 12,320 examples, this amounts to less than 2.5ms per example. The significance of this is highly context dependent. For instance, it may be insignificant relative to request latency or the time to encode an input and query in the case of serving a single request from an end user.

We also note that fine tuning all BERT layers with an adapter size of 64 yields \textit{only marginally} better results than fine tuning BERT alone, achieving an F1 score of 76.3\% relative to 75.8\%.

\end{multicols}
\begin{table}[t!]
    \centering
    \label{tab:table-1}
\begin{tabular*}{\textwidth}{c @{\extracolsep{\fill}} @{}*{8}{c}@{}}
    \toprule
      \thead{Model}
        & \thead{Adapter\\Size}
        & \thead{No. Layers\\Trained}
        & \thead{No. Trainable\\Params}
        & \thead{EM} 
        & \thead{F1} 
        & \thead{Train\\Time (s)} 
        & \thead{Inference\\Time (s)}\\
    \midrule
      L12-A64 & 64 & 12 & 110,691,074 & 73.2 & 76.3 & 269 & 154.3 \\
      L0-A64 & 64 & 0 & 2,417,664 & 66.8 & 69.8 & 203 & 154.8\\
      L0-A768 & 768 & 0 & 28,388,354 & 72.0 & 75.0 & 253 & 188.8 \\
    \bottomrule
\end{tabular*}
\caption{SQuAD2.0 results with variable adapter size and transformer layer freezing.}
\end{table}
\begin{multicols}{2}

\subsection{Context-Aware CNN}

Lastly, we explore context-aware CNNs as a final layer after BERT on the SQuAD2.0 task. The advantage of the downward projection to a single context vector as used in \cite{shen2017learning} is apparent in terms of performance relative to training time and number of trainable parameters. CACNN are clearly parameter inefficient without vector contextualization. Inference time, too, is negatively impacted by lack of context vectorization. We even see a significant deterioration in inference time with context vectorization as the number of per sample feature maps increases to 200s from 145s (base BERT model), representing about an 80\% increase in inference time. While this still only amounts to a 9.5ms increase per example, this may be significant for some online or batch processing production systems.

We also observe superior parameter efficiency of naïvely adding adapters relative to even the best performing CACNN. Ultimately, this suggests that even \textit{context-aware} CNNs have limited contextualization capacity relative to transformers. While CACNNs may be able to capture some contextual information, this is more precisely captured through the key-query-value mechanism of transformers. During a convolution operation, a filter may be able to contextualize adjacent words, and subsequently the reduction of the feature maps to a context vector may be able to capture the most salient features across a sequence of word vectors, but, in contrast to transformers, it has no mechanism to capture connections between individual words in different parts of a sentence.

\end{multicols}
\begin{table}[h!]
    \centering
    \label{tab:table-1}
\begin{tabular*}{\textwidth}{c @{\extracolsep{\fill}} @{}*{9}{c}@{}}
    \toprule
      \thead{Model}
        & \thead{Context\\Vector}
        & \thead{No. Feature\\Maps per\\example}
        & \thead{No. Layers\\Trained}
        & \thead{No. Trainable\\Params}
        & \thead{EM} 
        & \thead{F1} 
        & \thead{Train\\Time (s)} 
        & \thead{Inference\\Time (s)}\\
    \midrule
      L0-CACNNv120 & Y & 20 & 0 & 347,212 & 56.4 & 58.4 & 232 & 166.1 \\
      L0-CACNNv150 & Y & 50 & 0 & 838,824 & 57.7 & 60.1 & 246 & 176.2 \\
      L0-CACNNv200 & Y & 200 & 0 & 1,760,724 & 58.3 & 61.2 & 421 & 261.5 \\
      L0-CACNN4 & N & 4 & 0 & 141,603,852 & 57.3 & 60.0 & 481 & 243.7 \\
    \bottomrule
\end{tabular*}
\caption{SQuAD2.0 results with variable adapter size and transformer layer freezing.}
\end{table}
\begin{multicols}{2}

\section{Conclusions}

In these sets of experiments, we explored performance of adapter module and context-aware CNNs augmentation of BERT. Accounting for efficiency in terms of F1 score relative to trainable parameters, we found there to be some merit to using adapter modules to reduce the number of trainable parameters, while maintaining scores on the SQuAD2.0 task competitive with fine tuned BERT models. However, this benefit did not necessarily translate to reduced training time and of course adds to the inference time. The context-aware convolutional filters, on the other hand, failed to approach the F1 performance of the adapter modules, while also taking more time to train. Freezing the first six layers of the BERT model provides a simple and viable approach to reduce the total number of trainable parameters and training time without affecting the final model size and thus inference time.

\subsection{Acknowledgments}

The Lambda Labs workstation used to run these experiments was generously provided by Rigetti Computing. We would also like to acknowledge instructors from the UC Berkeley's Master in Data Science (MIDs) program for their guidance - Joachim Rahmfield, Alberto Todeschini, and Hossein Vahabi.

\newpage
\bibliography{bibliography.bib}
\bibliographystyle{ieeetr}
\end{multicols}

\newpage
\section{Appendix I: Figures}
\subsection{Appendix 1a}
\begin{figure}[h!]
  \includegraphics[width=\linewidth]{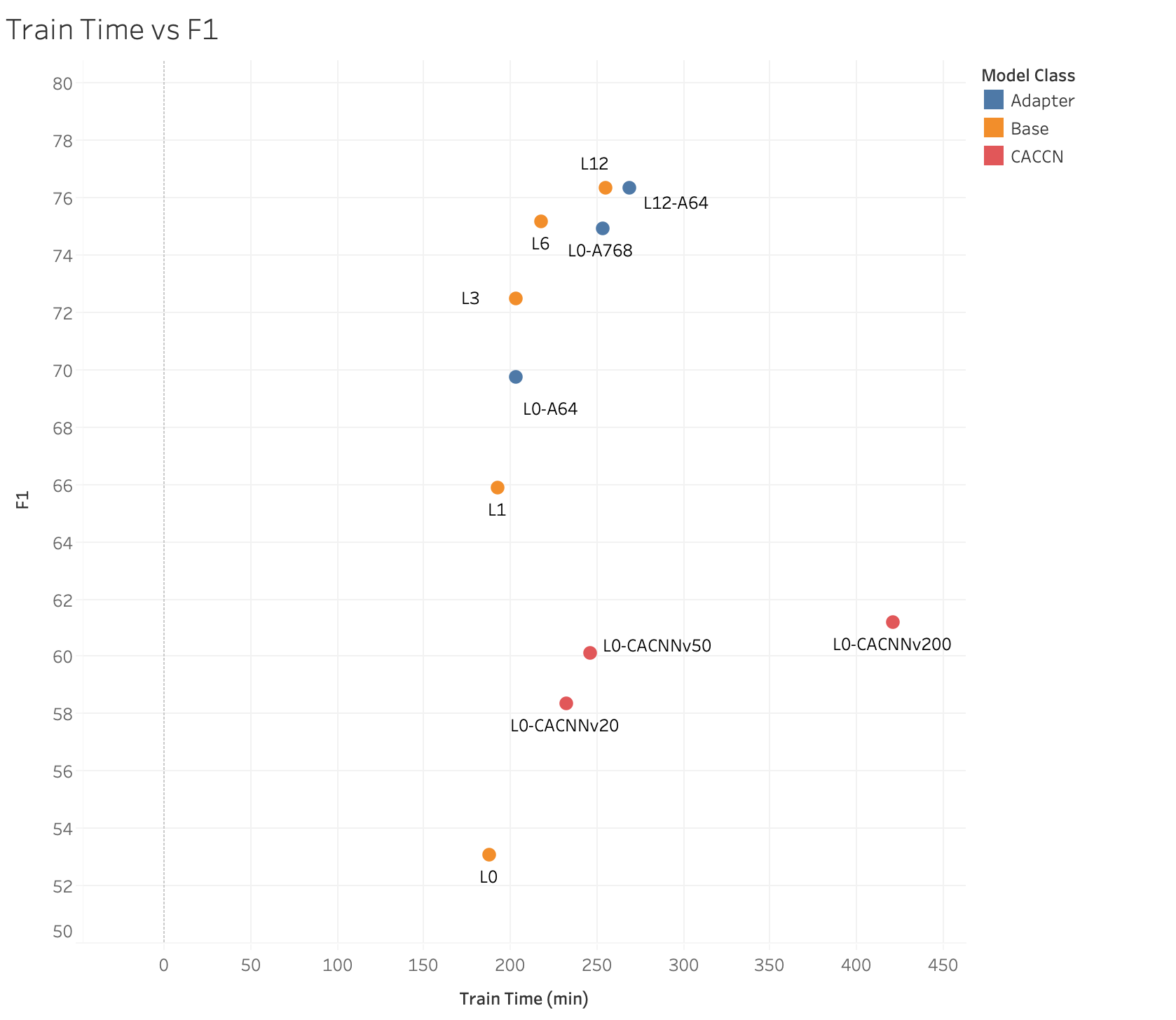}
  \caption{Time to train each model on SQuADv2.0 train dataset (132,300 examples) for three epocs vs corresponding F1 score on SQuAD2.0 test dataset.}
\end{figure}

\newpage
\subsection{Appendix 1b}
\begin{figure}[h!]
  \includegraphics[width=\linewidth]{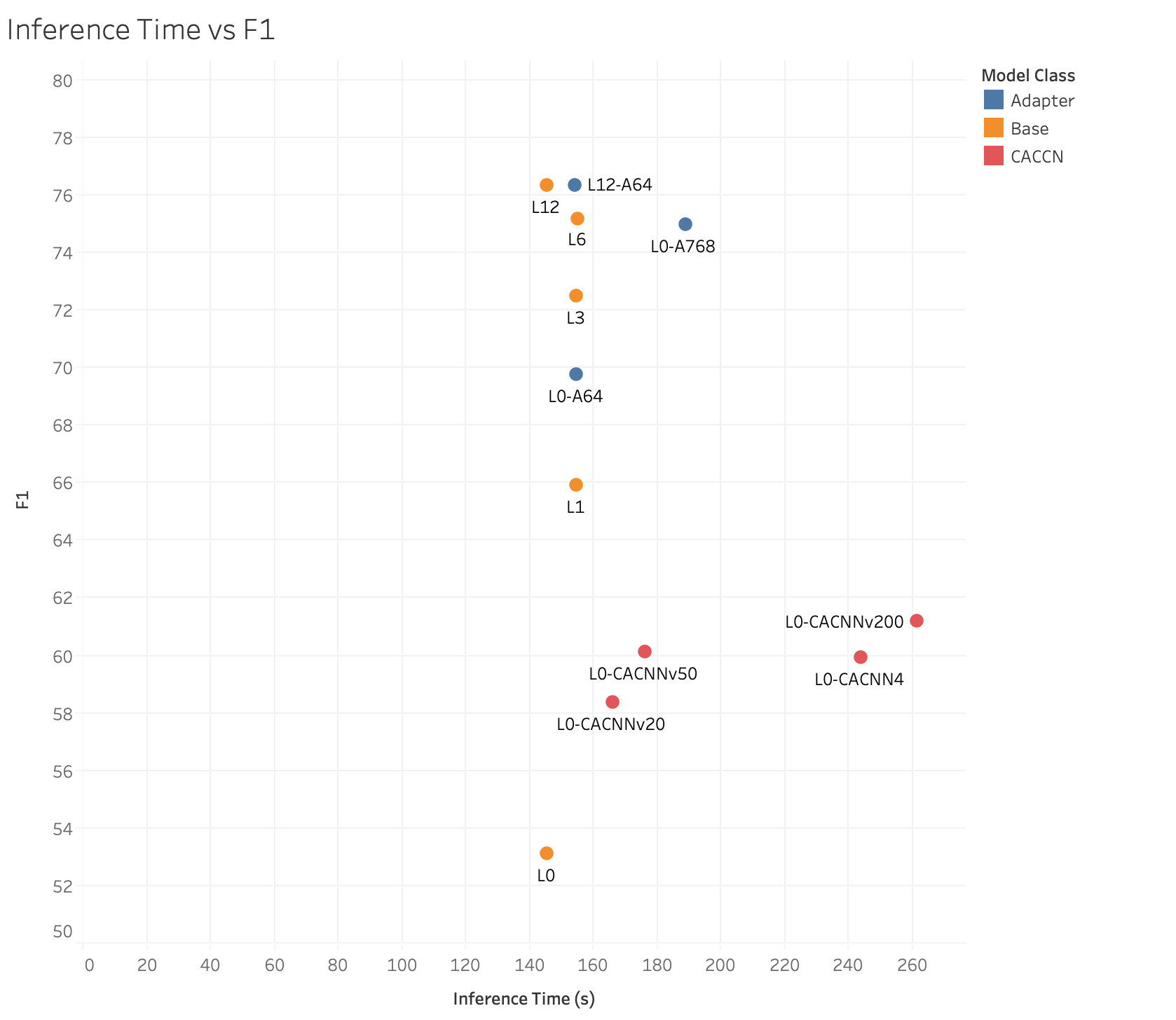}
  \caption{Time to run inference on pre-processed examples from SQuAD2.0 test dataset for each model vs corresponding F1 score.}
\end{figure}

\newpage
\subsection{Appendix 1c}
\begin{figure}[h!]
  \includegraphics[width=\linewidth]{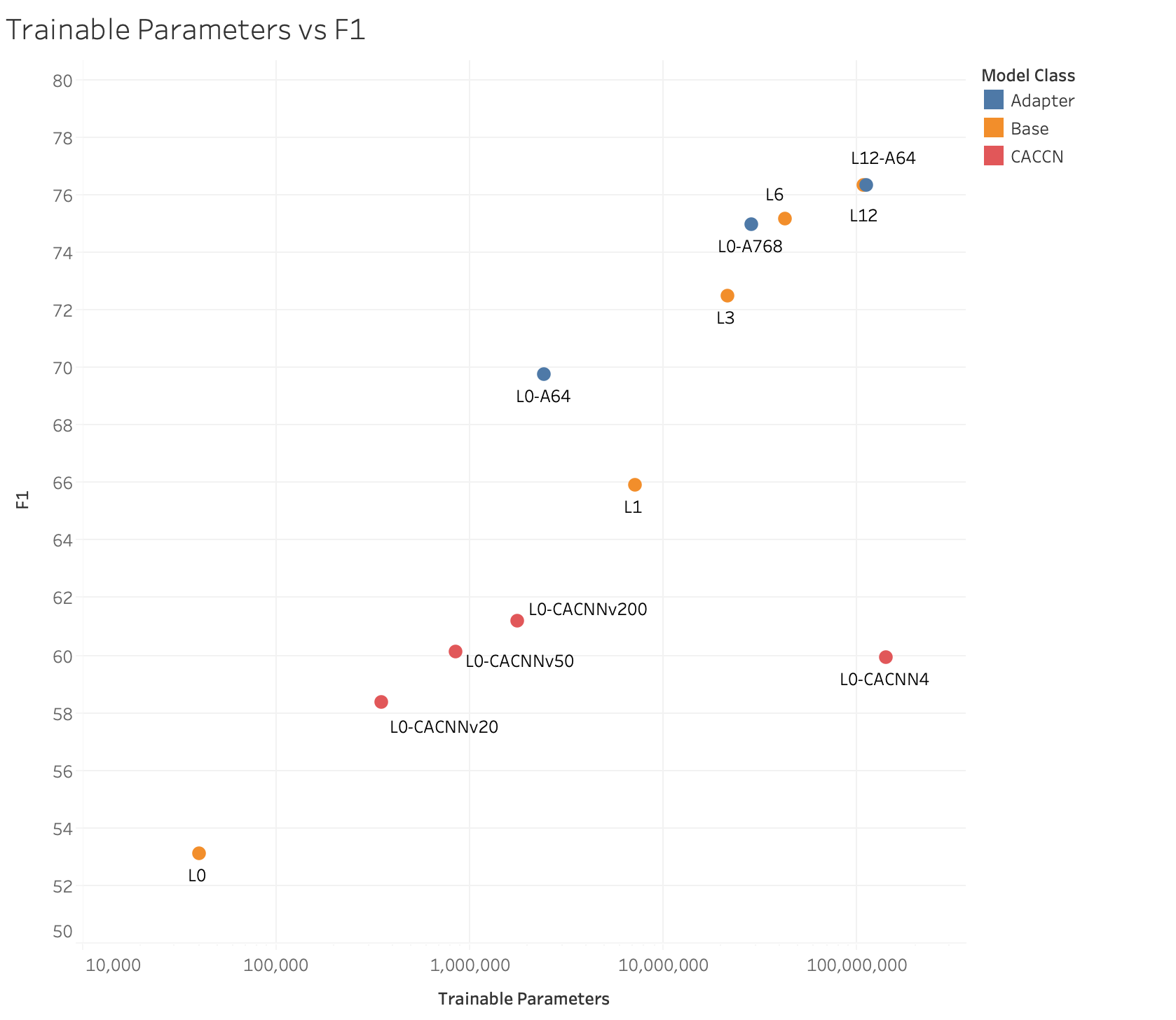}
  \caption{Number of parameters trained for each model vs corresponding F1 score on SQuAD2.0 test dataset.}
\end{figure}

\end{document}